\documentclass[11pt]{article}

\usepackage[final]{acl}

\usepackage{times}
\usepackage{latexsym}

\usepackage[T1]{fontenc}

\usepackage[utf8]{inputenc}

\usepackage{microtype}

\usepackage{inconsolata}

\usepackage{graphicx}
\usepackage{amssymb}
\usepackage{amsmath} 
\usepackage{todonotes} 
\usepackage{algorithm}
\usepackage{algorithmic}
\usepackage{multirow}
\newcommand{\ourM}{CAGenMol}
\title{\ourM{}: Condition-Aware Diffusion Language Model for Goal-Directed Molecular Generation}

\author{
 \textbf{Yanting LI\textsuperscript{1,}\thanks{These authors contributed equally to this work.}},
 \textbf{Zhuoyang JIANG\textsuperscript{1}\footnotemark[1]},
 \textbf{Enyan DAI\textsuperscript{1}},
 \textbf{Lei WANG\textsuperscript{2}},
 \\
 \textbf{Wen-Cai Ye\textsuperscript{2}},
 \textbf{Li LIU\textsuperscript{1}},
\\
 \textsuperscript{1} The Hong Kong University of Science and Technology (Guangzhou),
 \\
 \textsuperscript{2}Jinan University, Guangzhou
\\
 \small{
   \textbf{Correspondence:} \href{mailto:avrillliu@hkust-gz.edu.cn}{avrillliu@hkust-gz.edu.cn}
 }
}

\begin{document}
\maketitle
\begin{abstract}
Goal-directed molecular generation requires satisfying heterogeneous constraints such as protein--ligand compatibility and multi-objective drug-like properties, yet existing methods often optimize these constraints in isolation, failing to reconcile conflicting objectives (e.g., affinity vs. safety), and struggle to navigate the non-differentiable chemical space without compromising structural validity. 
To address these challenges, we propose \ourM{}, a condition-aware discrete diffusion framework over molecular sequences that formulates molecular design as conditional denoising guided by heterogeneous structural and property signals. 
By coupling discrete diffusion with reinforcement learning, the model aligns the generation trajectory with non-differentiable objectives while preserving chemical validity and diversity. 
The non-autoregressive nature of diffusion language model further enables iterative refinement of molecular fragments at inference time. 
Experiments on structure-conditioned, property-conditioned, and dual-conditioned benchmarks demonstrate consistent improvements over state-of-the-art methods in binding affinity, drug-likeness, and success rate, highlighting the effectiveness of our framework.
The code is available at \url{https://github.com/Lee612-1/CAGenMol}.
\end{abstract}

\begin{figure*}[t]
    \centering
    \includegraphics[width=\textwidth]{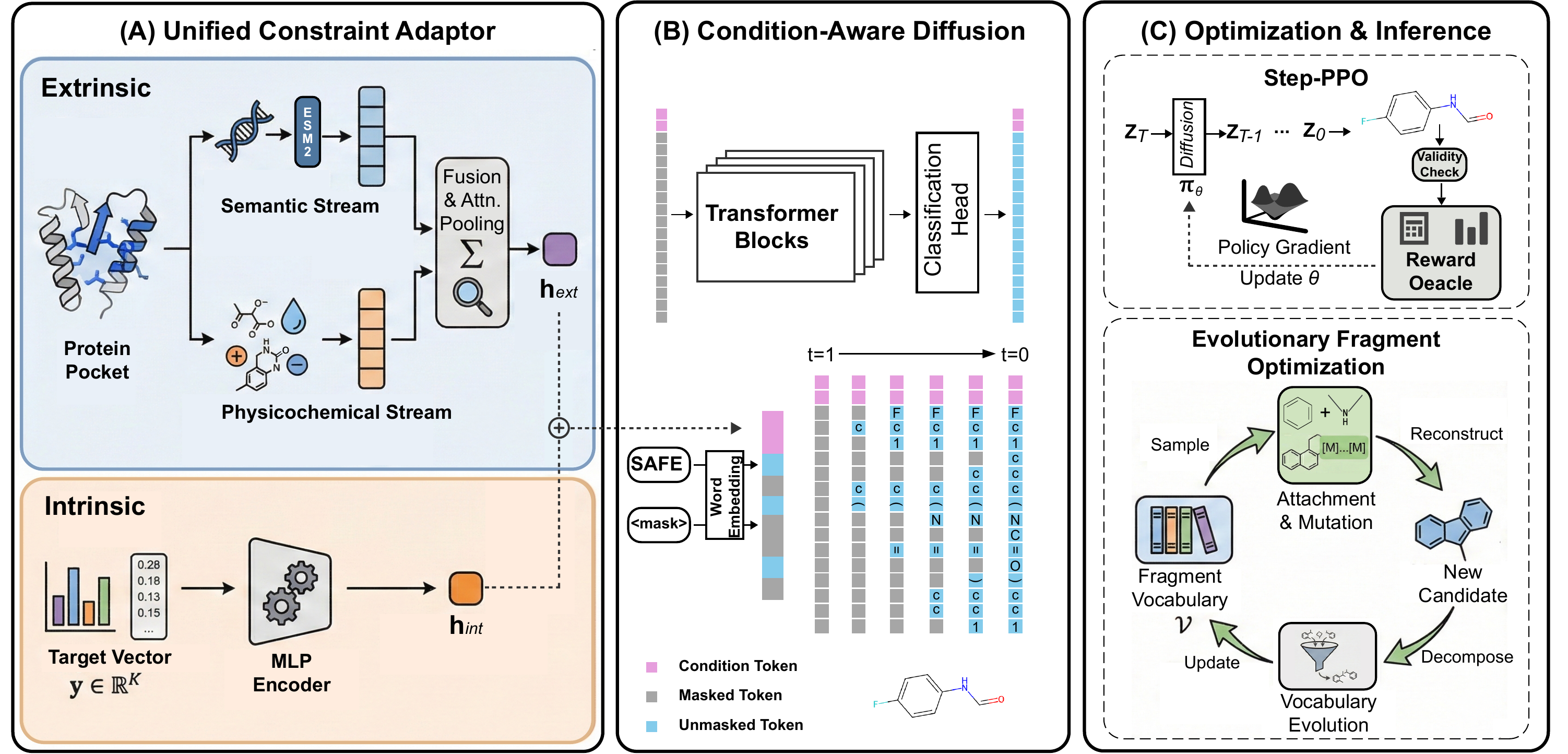}
    \caption{Overview of \ourM{}. UCA encodes either protein-pocket structure or target properties, which guides a condition-aware masked diffusion model on SAFE sequences. The model is further trained with Step-PPO, and perform inference with EFO refinement.}
    \label{fig:main}
\end{figure*}

\section{Introduction}

The discovery of novel small-molecule therapeutics is a cornerstone of modern medicine~\cite{dimasi2016innovation, hughes2011principles}. However, a clinical candidate must simultaneously satisfying diverse application settings and their corresponding constraints~\cite{polishchuk2013estimation, ferreira2019admet}. This multi-objective nature makes the search space vast and discontinuous, rendering traditional trial-and-error approaches inefficient. Consequently, generative models have emerged as a promising paradigm to accelerate goal-directed molecular design.

Despite their promise, existing approaches typically compromise between structural precision and optimization flexibility. Structure-Based Drug Design (SBDD) methods directly model 3D atomic interactions to ensure high binding affinity~\cite{peng2022pocket2mol, guan20233d}. However, they rely on computationally expensive 3D representing and struggle to optimize non-geometric pharmacological properties. Conversely, sequence-based optimization methods treat molecular generation as a black-box search~\cite{loeffler2024reinvent, zhou2019optimization}. While flexible in defining objectives, they often lack structural priors, leading to chemically invalid results~\cite{renz2019failure}. Therefore, a unified framework that reconciles structural perception with robust multi-objective optimization remains an open challenge.

From a modeling perspective, 1D sequence representations (e.g., SMILES) offer a computationally efficient alternative to 3D conformations. However, the dominant Autoregressive (AR) models~\cite{bagal2021molgpt, noutahi2024gotta} generate tokens in a rigid left-to-right order. This mechanism fundamentally limits their ability to incorporate global structural contexts or perform local refinement, making them fragile when coupled with aggressive RL policy.
To address these limitations, we identify Discrete Diffusion Language Models (DLMs) as a superior substrate for goal-directed generation. DLMs generate molecules via a non-autoregressive iterative denoising process~\cite{austin2021structured, sahoo2024simple}. This paradigm offers two implicit advantages: (1) Global Visibility, which allows the model to attend to conditioning signals simultaneously across the entire sequence; and (2) Iterative Editability, which permits fine-grained structural corrections during generation. Despite these potentials, how to effectively ground DLMs on complex biological constraints and optimize them via reinforcement learning in discrete space remains underexplored.

In this work, we propose \ourM{}, a unified framework that synergizes condition-aware discrete diffusion with reinforcement learning to bridge the gap between biological constraints and chemical space.
First, to handle heterogeneous inputs, we introduce a \textbf{Unified Constraint Adaptor (UCA)} that projects diverse signals into a shared latent space, serving as a persistent semantic anchor for the diffusion backbone.
Building on diffusion process, we develop \textbf{Step-wise Proximal Policy Optimization (Step-PPO)}. Unlike token-level RL on AR model, Step-PPO leverages the iterative nature of diffusion to perform fine-grained credit assignment at each denoising step, ensuring precise alignment with complex objectives without collapsing the generative prior.
Finally, to surpass the limitations of one-shot sampling, we propose \textbf{Evolutionary Fragment Optimization (EFO)}, an inference-time refinement strategy that exploits the editing flexibility of our model to perform gradient-free hill-climbing on generated candidates.


Our contributions are summarized as follows:
(1) We formulate goal-directed molecular generation as a conditional discrete diffusion problem, providing a unified modeling perspective that naturally accommodates heterogeneous structural and property constraints.
(2) We establish a diffusion-aware optimization framework that enables step-level policy learning within the denoising process, allowing effective alignment with complex objectives in discrete chemical space.
(3) We propose a principled inference-time refinement mechanism that leverages the flexibility of non-autoregressive diffusion models to iteratively improve generated molecules while preserving diversity.
(4) Comprehensive experiments across multiple conditional generation settings validate the effectiveness and robustness of the proposed framework.

\section{Related Work}

\textbf{Diffusion Language Models}
Transformer architectures~\cite{vaswani2017attention} have revolutionized sequence modeling. While autoregressive models such as GPT~\cite{radford2018improving} achieve strong performance, their left-to-right decoding results in slow inference and limited editability. Discrete Diffusion Language Models (DLMs) mitigate these issues through non-autoregressive generation. By introducing diffusion processes over discrete token spaces~\cite{austin2021structured,lou2023discrete,sahoo2024simple}, DLMs enable faster generation, greater diversity, and flexible token ordering.
Recent work such as LLaDA-V~\cite{you2025llada} extends DLMs to multimodal settings; however, their application to scientific domains requiring precise conditioning on heterogeneous biological signals remains largely unexplored.

\textbf{Structure-Based Drug Design}
Structure-Based Drug Design(SBDD) aims to generate ligands that specifically bind to a target protein pocket. Existing methods~\cite{peng2022pocket2mol,guan20233d, jain2023multi} typically model the joint 3D atomic distribution, with recent work~\cite{qu2024molcraft,zhang2025molchord} further incorporating latent diffusion or geometric vector quantization.
Despite their success, these approaches face two key limitations: they rely on computationally expensive explicit 3D representations that assume static protein structures, and they primarily optimize geometric or binding criteria while neglecting broader drug-like properties such as ADMET and synthetic accessibility.

\textbf{Optimization-Based Molecular Generation}
Goal-directed molecular generation is commonly formulated as an optimization problem, with Reinforcement Learning (RL)~\cite{loeffler2024reinvent,zhou2019optimization, wang2024efficient}and Genetic Algorithms (GA)~\cite{yoshikawa2018population,spiegel2020autogrow4,jensen2019graph} as the dominant paradigms which respectively optimize generative policies toward target objectives and evolve molecular populations via mutation and crossover.
While both paradigms support flexible objective definitions and thus offer strong generality, they typically treat chemical space as a black box, relying on weak generative priors and sparse reward signals. As a result, RL-based methods are prone to reward hacking and mode collapse, whereas GA-based approaches often generate chemically invalid molecules.

\section{Preliminary}
\subsection{Problem Definition}\label{sec:problem_definition}
We formulate goal-directed molecular generation as a conditional sequence generation task, aiming to learn a distribution $p_\phi(\mathcal{S}\mid\mathbf{c_s, c_p})$ over molecular sequences $\mathcal{S}=[s_1,\dots,s_L]$, where $\mathbf{c_s, c_p}$ are:

\textbf{Extrinsic Structural Condition}. $\mathbf{c}_s$ represents a 3D protein pocket $\mathcal{P}={(\mathbf{r}_i,\mathbf{v}_i)}_{i=1}^N$, where $\mathbf{r}_i$ and $\mathbf{v}_i$ denote atomic coordinates and chemical features. The goal is to generate ligands that bind $\mathcal{P}$ with high affinity.

\textbf{Intrinsic Property Condition.} $\mathbf{c}_p$ specifies a $K$-dimensional target property vector $\mathbf{y}\in\mathbb{R}^K$, and the objective is to generate molecules whose properties match $\mathbf{y}$.
In both cases, $p\phi$ must generate valid molecular structures that satisfy the given condition $\mathbf{c}$.

\subsection{Diffusion Language Model}
\label{sec:background_mdlm}

Masked diffusion models~\cite{sahoo2024simple,shi2024simplified} are an effective class of diffusion language models. We follow MDLM~\cite{sahoo2024simple} to define the masked diffusion process.

Let $\pmb{x}$ be a length-$L$ sequence, where each token $\pmb{x}^{l}$ is a one-hot vector over $K$ categories, satisfying $\pmb{x}^{l}_i \in \{0,1\}^K$ and $\sum_{i=1}^{K} \pmb{x}^{l}_i = 1$. The $K$-th category corresponds to the masking token $\mathbf{m}$, with $\mathbf{m}_K = 1$. We denote by $\mathrm{Cat}(\cdot;\pmb{\pi})$ a categorical distribution with parameter $\pmb{\pi} \in \Delta^K$.

The forward process gradually replaces clean tokens with the mask according to
\begin{equation}
\resizebox{0.8\columnwidth}{!}{$
    q(\pmb{z}^{l}_t \mid \pmb{x}^{l})
    = \mathrm{Cat}\!\left(
    \pmb{z}^{l}_t;\,
    \alpha_t \pmb{x}^{l} + (1-\alpha_t)\mathbf{m}
    \right),
    \label{eq:fwd_masked}
$}
\end{equation}
where $\pmb{z}^{l}_t$ denotes the $l$-th token at time $t \in [0,1]$, and $\alpha_t$ is a monotonically decreasing masking schedule with $\alpha_0 = 1$ and $\alpha_1 = 0$. At $t=1$, all tokens are masked.

The reverse process recovers less-masked sequences from more-masked ones. For $s<t$, the reverse transition $p_\theta(\pmb{z}^{l}_s \mid \pmb{z}^{l}_t)$ is defined as
\begin{equation}
    \resizebox{\hsize}{!}{$
    \begin{aligned}
        \begin{cases}
            \mathrm{Cat}(\pmb{z}^{l}_s; \pmb{z}^{l}_t),
            & \pmb{z}^{l}_t \neq \mathbf{m}, \\
            \mathrm{Cat}\!\left(
            \pmb{z}^{l}_s;\,
            \dfrac{(1-\alpha_s)\mathbf{m} + (\alpha_s - \alpha_t)\,
            \pmb{x}^{l}_\theta(\pmb{z}_t, t)}
            {1-\alpha_t}
            \right),
            & \pmb{z}^{l}_t = \mathbf{m},
        \end{cases}
    \end{aligned}
    $}
    \label{eq:rev_masked}
\end{equation}
where $\pmb{x}_\theta(\pmb{z}_t, t)$ is a denoising network predicting the clean-token distributions. This formulation preserves already unmasked tokens.

\subsection{SAFE and Base Model}
We adopt SAFE~\cite{noutahi2024gotta} to align with the non-autoregressive nature of diffusion and ensure structural stability during RL exploration. Unlike SMILES~\cite{weininger1988smiles,krenn2020self}, SAFE's fragment-based representation imposes strong chemical priors, preventing local invalidity even under aggressive optimization. We further initialize our framework with the pre-trained GenMol~\cite{lee2025genmol} backbone to inherit learned chemical distributions, allowing the model to focus exclusively on condition alignment rather than learning basic validity.

\section{Methodology}
As illustrated in Figure \ref{fig:main}, we present a unified framework \ourM{} for goal-directed molecular generation that synergizes condition-aware discrete diffusion with reinforcement learning. The framework is designed to explicitly align molecular generation with complex biochemical objectives beyond pure data distribution modeling.

\subsection{Model Architecture.}

The core design philosophy of \ourM{} is to bridge the modality gap between heterogeneous biological constraints and discrete chemical space. To achieve this, we structurally decouple constraint perception from molecular reasoning. The architecture is composed of two synergistic modules: a \textbf{Unified Constraint Adaptor (UCA)} that projects diverse signals (e.g., 3D geometric pockets or 1D property vectors) into a shared latent semantic space, and a \textbf{Condition-Aware Diffusion Backbone} that utilizes these unified representations to bias the discrete denoising trajectory. This design allows a single generative framework to flexibly adapt to both extrinsic structural environments and intrinsic property requirements.
\subsubsection{Unified Constraint Adaptor}
\label{sec:condition_adaptor}

To map heterogeneous constraint signals into the shared latent space of the diffusion backbone with dimension $D$, 
UCA acts as a learnable interface that translates biological and chemical constraints into a unified latent guidance representation.

\textbf{Extrinsic Constraint: Structure Adaptation.}
A protein pocket defines the external physicochemical environment that constrains the feasible geometric space and interaction patterns of a ligand~\cite{koshland1958application, schneider2005computer}. 
To encode this extrinsic constraint, we propose a dual-stream encoding strategy to bridge the gap between implicit evolutionary semantics and explicit surface chemistry. While protein language models capture long-range dependencies, they often lack the granularity required for precise interaction matching. Therefore, we augment semantic features with explicit physicochemical descriptors:

\textbf{(1) Semantic Stream.} We extract residue-level embeddings $\mathbf{H}_{esm} \in \mathbb{R}^{L_{pocket} \times 1280}$ using the pre-trained ESM-2~\cite{lin2023evolutionary}, leveraging its evolutionary knowledge to characterize the pocket's biological context.

\textbf{(2) Physicochemical Stream.} To explicitly guide interaction matching, we compute a 5-dimensional feature vector $\mathbf{h}_{phys}$ for each residue (e.g., charge, hydropathy, H-bond potential). These are stacked to form $\mathbf{H}_{phys} \in \mathbb{R}^{L_{pocket} \times 5}$, ensuring the model attends to key binding determinants.

Both streams are projected into the shared model dimension $D$ via independent MLPs to align their semantic manifolds:
\begin{equation}
    \resizebox{0.8\columnwidth}{!}{$
    \tilde{\mathbf{H}}_{esm} = \mathrm{MLP}_{esm}(\mathbf{H}_{esm}^{pocket}) \in \mathbb{R}^{L_{pocket} \times D}, $}
\end{equation}

\begin{equation}
    \resizebox{0.8\columnwidth}{!}{$
    \tilde{\mathbf{H}}_{phys} = \mathrm{MLP}_{phys}(\mathbf{H}_{phys}^{pocket}) \in \mathbb{R}^{L_{pocket} \times D}.
    $}
\end{equation}
The projected features are then fused by element-wise summation to yield the unified pocket representation $\mathbf{H}_{fused} = \tilde{\mathbf{H}}_{esm} + \tilde{\mathbf{H}}_{phys}$.

To identify key binding residues without relying on explicit 3D coordinates, we employ a Linear Attention Pooling mechanism. By computing a learnable importance score for each residue, the model autonomously learns to focus on functional hotspots in the pocket. Specifically, the attention weights $\boldsymbol{\alpha} \in \mathbb{R}^{L_{pocket} \times 1}$are computed as:
\begin{equation}
    \resizebox{0.7\columnwidth}{!}{$
    \boldsymbol{\alpha} = softmax(\mathrm{MLP}_{attn}(\mathbf{H}_{fused})).
    $}
\end{equation}
The final extrinsic condition token is then obtained as a capability-weighted sum, ensuring that the guidance signal is dominated by the most pharmacologically relevant residues:
\begin{equation}
    \resizebox{0.7\columnwidth}{!}{$
    \mathbf{h}_{ext}
    =
    \sum_{i=1}^{L_{pocket}}
    \alpha_i \mathbf{H}_{fused}^{(i)}
    \in \mathbb{R}^{1 \times D}.
    $}
\end{equation}

\textbf{Intrinsic Constraint: Property Adaptation.}
Beyond structural fit, drug candidates must satisfy intrinsic property constraints $\mathbf{y} \in \mathbb{R}^{K}$ (e.g., ADMET). To translate these scalar values into high-dimensional guidance signals compatible with the diffusion sequence, the UCA projects $\mathbf{y}$ into the latent space via a learnable mapping $\mathrm{MLP}_{int}$, producing a property-conditioned token $\mathbf{h}_{int} \in \mathbb{R}^{1 \times D}$. This enables the diffusion model to interpret numerical properties as semantic prompts.

\subsection{Condition-Aware Diffusion Backbone}

We adapt the BERT-based GenMol~\cite{lee2025genmol} architecture for conditional generation. Instead of introducing heavy cross-attention modules which require training from scratch, we propose a parameter-efficient \textbf{Prompt-based Conditional Denoising} strategy. This approach treats the condition vector not merely as an input, but as a semantic prompt that prefixes the molecular sequence.

Formally, we construct the input embedding $\mathbf{H}_t$ at time step $t$ by prepending the condition token $\mathbf{h_c}$ (derived from UCA) to the noisy molecular embeddings:
\begin{equation}
    \resizebox{0.8\columnwidth}{!}{$
    \mathbf{H}_t =
    \big[
    \mathbf{h_c},\,
    \mathrm{Embed}(\pmb{z}_t^1),\,
    \dots,\,
    \mathrm{Embed}(\pmb{z}_t^L)
    \big].
    $}
    \end{equation}

This design leverages the \textbf{global broadcasting capability} of bidirectional self-attention. Since $\mathbf{h_c}$ is visible to every molecular token at every layer, it serves as a \textbf{persistent semantic anchor} throughout the diffusion process. Even when the molecular sequence $\pmb{z}_t$ is heavily corrupted by mask tokens (in the forward process), the unmasked $\mathbf{h_c}$ provides a stable reference signal. This allows the model to effectively bias the denoising distribution $p_\theta(\pmb{x}_0 \mid \pmb{z}_t, \mathbf{h_c})$ toward the target chemical manifold without disrupting the pre-trained structural priors.

\subsection{Training and Inference Pipeline.}
\ourM{} is optimized and deployed following a three-stage paradigm. First, the model is trained via supervised learning with a discrete diffusion objective, which provides a stable initialization for subsequent optimization. Second, we introduce a step-wise Proximal Policy Optimization (Step-PPO) algorithm to further steer the generation process toward task-specific objectives. Finally, during inference, an Evolutionary Fragment Optimization (EFO) procedure is applied to iteratively refine and improve the generated molecular candidates.

\subsubsection{Supervised Learning}
We first train \ourM{} via supervised learning to adapt the unconditional backbone to conditioning signals. Following MDLM~\cite{sahoo2024simple}, we optimize a continuous-time approximation of the negative evidence lower bound (NELBO), which acts as a time-weighted masked language modeling objective over the molecular tokens (details in Appendix~\ref{app:obj}). This stage establishes a stable, condition-aware initialization for subsequent RL alignment.

\subsubsection{Step-wise PPO for Diffusion Language Model}
\label{sec:step-ppo}
While the supervised stage establishes a chemical prior, it prioritizes likelihood over functional desirability, often failing to explore high-reward regions defined by non-differentiable oracles (e.g., docking). Thus, we propose \textbf{Step-wise Proximal Policy Optimization (Step-PPO)}. 
Unlike traditional trajectory-level RL, Step-PPO reformulates discrete diffusion as a fine-grained Markov Decision Process (MDP), enabling precise alignment with complex objectives while preserving generative coherence.

\textbf{Algorithm.} We can interpret the reverse diffusion process as a MDP, which naturally arises from the iterative denoising formulation. At each diffusion time step $t$, the state $s_t = \pmb{z}_t$ corresponds to the partially masked molecular sequence, and the policy $\pi_\theta$ (parameterized by the diffusion model) defines a distribution over actions $a_t \sim \pi_\theta(\cdot \mid s_t)$,
where an action $a_t$ corresponds to selecting categorical tokens to replace the masked positions in $\pmb{z}_t$ during the reverse transition from $t$ to $t-1$. This formulation satisfies the Markov property, as each denoising step depends only on the current sequence state.

Unlike prior diffusion-based RL approaches that treat the entire denoising trajectory as a single action \cite{zhao2025d1, shao2024deepseekmath, yang2025taming}, we apply policy optimization \cite{schulman2017proximal} at each diffusion step. 
As intermediate masked states lack defined chemical properties, we formulate the task as a \textbf{sparse reward problem}, maximizing the terminal reward $R$ evaluated solely at $t=0$.

To stabilize policy updates, we adopt the clipped surrogate objective from PPO~\cite{schulman2017proximal}. For a specific diffusion step $t$ involving an action $a_t$, the loss function is defined as:
\begin{equation}
\resizebox{1.0\columnwidth}{!}{$
\mathcal{L}_{\text{step}}^{(t)}(\theta) = - \mathbb{E}_{\pi_{\theta_{\text{old}}}} \left[ \min \left( r_t(\theta) \hat{A}_t, \, \text{clip}(r_t(\theta), 1-\epsilon, 1+\epsilon) \hat{A}_t \right) \right]
$}
\end{equation}
where $r_t(\theta) = \frac{\pi_\theta(a_t \mid s_t)}{\pi_{\theta_{\text{old}}}(a_t \mid s_t)}$ denotes the probability ratio between the current and behavioral policies.

To efficiently estimate the signal without an additional value network, we compute the advantage $\hat{A}_t$ using batch-level reward statistics:
\begin{equation}
\resizebox{0.3\columnwidth}{!}{$
\hat{A}_t = \frac{R - \mu_{\mathcal{B}}}{\sigma_{\mathcal{B}} + \epsilon},
$}
\end{equation}
where $R$ represents the terminal reward of the current trajectory, while $\mu_{\mathcal{B}}$ and $\sigma_{\mathcal{B}}$ denote the mean and standard deviation of valid rewards within the sampling batch $\mathcal{B}$.

Moreover, rewards are only well-defined for chemically valid molecules. So we introduce a Validity Mask $\mathcal{M}_{valid} \in \{0,1\}$, which activates policy updates only for valid trajectories and effective denoising steps. 
To ensure stability, we introduce a mask $\mathcal{M}_{valid} \in \{0,1\}$ to restrict updates to chemically valid trajectories. Given the pre-trained backbone's high validity rate ($>90\%$), this mechanism imposes negligible overhead and serves effectively as a noise filter, preventing undefined reward signals from corrupting gradient estimation.

The final optimization objective over a batch of size $B$ is given by:
\begin{equation}
\resizebox{0.8\columnwidth}{!}{$
\mathcal{L}_{\mathrm{batch}}
=
\frac{
\sum_{b=1}^{B}
\mathcal{M}_{valid}^{(b)}
\left(
\sum_{t}
\mathcal{L}_{\mathrm{step}}^{(b,t)}
+
\beta \mathcal{L}_{\mathrm{ent}}^{(b,t)}
\right)
}{
\sum_{b=1}^{B}
\mathcal{M}_{valid}^{(b)}
},
$}
\end{equation}
where $\mathcal{L}_{\mathrm{ent}}$ denotes the entropy regularization term. This design ensures that policy updates are driven exclusively by valid molecular trajectories that satisfy the imposed conditions.

\textbf{Reward Design.} While the optimization algorithm is generic, the reward function encodes task-specific objectives. We design task-specific terminal reward functions for structure-conditioned and property-conditioned molecular generation, both evaluated on the fully denoised molecule at the final step $t=0$.

For structure-conditioned generation, the primary objective is to optimize the protein-ligand interaction strength. We quantify this using the standard Vina score ($S_{\mathrm{dock}}$) computed by AutoDock Vina~\cite{eberhardt2021autodock}, which approximates the Gibbs free energy of binding ($\Delta G$) in kcal/mol.
Since raw docking scores are unbounded and highly target-dependent, we introduce a reference affinity threshold $S_{\mathrm{ref}}$ (determined empirically from the reference ligand's affinity, see Sec.~\ref{sec:exp_struct}). We then define the affinity margin $\Delta = S_{\mathrm{ref}} - S_{\mathrm{dock}}$ to measure relative improvement.
The structure-based reward is defined as:
\begin{equation}
\resizebox{0.85\columnwidth}{!}{$
R_{\mathrm{struct}}
=
\mathrm{sign}(\Delta)\cdot \Delta^{2}
+
\lambda_1 \cdot \mathrm{QED}
+
\lambda_2 \cdot \mathrm{SA},
$}
\label{eq:reward struct}
\end{equation}
where $\mathrm{QED}$ \cite{bickerton2012quantifying} and $\mathrm{SA}$ \cite{ertl2009estimation} encourage drug-likeness and synthetic accessibility. $\lambda_1$ and $\lambda_2$ are balancing coefficients used to normalize the scales of different objectives. This quadratic formulation implicitly prioritizes the refinement of pharmaceutical properties when the binding affinity approaches the reference threshold.

For property-conditioned generation, let $\mathbf{y}_{\mathrm{tgt}} \in \mathbb{R}^{K}$ denote the target property vector and $\hat{\mathbf{y}} \in \mathbb{R}^{K}$ the predicted properties. 
To harmonize heterogeneous property scales into a bounded reward space $[0,1]$, we map the prediction error to a similarity score using a weighted Gaussian kernel:
\begin{equation}
\resizebox{0.8\columnwidth}{!}{$
R_{\mathrm{prop}}
=
\sum_{k=1}^{K}
\omega_k
\exp\!\left(
-\frac{(\hat{y}_k - y_{\mathrm{tgt},k})^2}{2\sigma_k^2}
\right).
$}
\end{equation}
To calibrate for heterogeneous scales and optimization difficulty, we derive $\sigma_k$ and $\omega_k$ from statistics of 1,000 molecules sampled from the initial policy $\pi_{\text{init}}$. Specifically, $\sigma_k$ is defined as the empirical standard deviation $\mathrm{Std}_{\mathcal{D}}(y_k)$ of these samples to normalize diverse numerical ranges. Meanwhile, the weight $\omega_k = \varepsilon_k / \sum_{j=1}^{K} \varepsilon_j$ is assigned proportional to the mean absolute error $\varepsilon_k = \mathbb{E}_{\mathcal{D}}[|\hat{y}^{(0)}_k - y_k|]$ on this batch, adaptively prioritizing properties that are initially harder to satisfy.

\subsubsection{Evolutionary Fragment Optimization}
\label{sec:inference}
To mitigate sampling stochasticity, we introduce \textbf{Evolutionary Fragment Optimization (EFO)} to perform gradient-free hill-climbing at inference time. EFO iteratively refines candidates by resampling masked substructures via the conditional diffusion backbone.
Formally, for a molecule $x$, we apply a mask to select $x_{\text{mask}} \subset x$ and sample a new candidate $x'$ conditioned on the remaining context:
\begin{equation}
x' \sim p_\theta(x_{\text{mask}} \mid x \setminus x_{\text{mask}}, \mathbf{h_c}).
\end{equation}
The fragment vocabulary $\mathcal{V}$ is dynamically updated by decomposing generated candidates and retaining the top-$K$ structures based on property scores $S(\cdot)$:
\begin{equation}
\mathcal{V}_{t+1} \leftarrow \text{TopK}\big( \mathcal{V}_t \cup \text{Decompose}(x'), \; S \big).
\end{equation}
This loop concentrates the search on high-value chemical regions (see Algorithm~\ref{alg:efo} in Appendix).

\section {Experiments}

We conduct three sets of experiments to evaluate \ourM{} under (i) structure-conditioned generation, (ii) multi-target property-conditioned generation and (iii) dual-conditioned generation. All training and evaluation are performed on a single \textbf{NVIDIA A800} GPU.

\subsection{Structure-Conditioned Generation}
\label{sec:exp_struct}

\begin{table*}[t]
\centering
{
\begin{tabular}{cccccccc}
\hline
Methods & Vina Dock ($\downarrow$) &High Affinity ($\uparrow$) & QED ($\uparrow$) & SA ($\uparrow$) & Diversity ($\uparrow$) & Success Rate ($\uparrow$)
\\ \hline 
Reference Set & -7.45 & - & 0.48 & 0.73 & - & 25.0\% \\ \hline
TargetDiff & -7.80 &58.1\%& 0.48 & 0.58 & 0.72 & 10.5\% \\
FLAG & -5.63 &-& 0.49 & 0.70 & 0.70 & 14.1\% \\
Pocket2Mol & -7.15 &48.4\%& 0.56 & 0.74 & 0.69 & 24.4\% \\
DecompDiff & -8.39 &64.4\%& 0.45 & 0.61 & 0.68 & 24.5\% \\
MolCRAFT & \textbf{-9.25} &59.1\%& 0.46 & 0.62 & 0.61 & 36.1\% \\ 
RGA + Vina & -8.01 &64.4\%& \underline{0.57} & 0.71 & 0.41 & 46.2\% \\ 
DecompOpt & \underline{-8.98} &73.5\%& 0.48 & 0.65 & 0.60 & 52.5\% \\
MOLCHORD & -8.59 &74.6\%& 0.56 & \underline{0.78} & 0.71 & 53.4\% \\
\hline
\ourM{} & -8.41 &\textbf{82.3\%}& \textbf{0.70} & \textbf{0.89} & \textbf{0.75} & \textbf{69.7\%} \\
\hline
\end{tabular}}
\caption{Comparison on the CrossDocked2020 benchmark. We report the average metrics across 100 test pockets. The best results are highlighted in \textbf{bold}, and the second best are \underline{underlined}.}
\label{tab:res1}
\end{table*}

\begin{table*}[h]
\centering
\resizebox{\textwidth}{!}{%
\begin{tabular}{lccccc}
\hline
\textbf{Variant} & \textbf{Vina Dock} ($\downarrow$) & \textbf{QED} ($\uparrow$) & \textbf{SA} ($\uparrow$) & \textbf{Diversity} ($\uparrow$) & \textbf{Success Rate} ($\uparrow$) \\ \hline
SFT (Base) & -6.47 & 0.53 & 0.77 & \textbf{0.88} & 14.3\% \\
SFT (w/o Attn.) & -6.60 & 0.55 & 0.76 & 0.86 & 17.5\% \\
SFT (w/o Phys.) & -6.55 & 0.54 & 0.78 & 0.87 & 15.8\% \\
SFT (Full UCA) & -6.61 & 0.58 & 0.77 & 0.86 & 19.2\% \\ \hline
\textbf{Full Model (Step-PPO)} & \textbf{-8.41} & \textbf{0.70} & \textbf{0.89} & 0.80 & \textbf{69.7\%} \\ \hline
\end{tabular}%
}
\caption{Ablation study of UCA components and RL training. All variants except the final one are trained solely with Supervised Fine-Tuning (SFT).}
\label{tab:ablation_components}
\end{table*}

\begin{table*}[t]
\centering
\resizebox{\textwidth}{!}{
\begin{tabular}{lcccccccc}
\hline
Model & 1IEP & 3EML & 3NY8 & 4RLU & 4UNN & 5MO4 & 7L11 & Avg \\
\hline
3DSBDD & -9.05$\pm$0.38 & -10.02$\pm$0.15 & -10.10$\pm$0.24 & -9.80$\pm$0.55 & -8.23$\pm$0.30 & -8.71$\pm$0.45 & -8.47$\pm$0.18 & -9.20 \\
AutoGrow4 & -13.23$\pm$0.11 & -13.03$\pm$0.09 & -11.70$\pm$0.00 & -11.20$\pm$0.00 & -11.14$\pm$0.12 & -10.38$\pm$0.27 & -8.84$\pm$0.33 & -11.36 \\
Pocket2Mol & -10.17$\pm$0.53 & -12.25$\pm$0.27 & -11.89$\pm$0.16 & -10.57$\pm$0.12 & -12.20$\pm$0.34 & -10.07$\pm$0.62 & -9.74$\pm$0.38 & -10.98 \\
PocketFlow & -12.49$\pm$0.70 & -9.25$\pm$0.29 & -8.56$\pm$0.35 & -9.65$\pm$0.25 & -7.90$\pm$0.78 & -7.80$\pm$0.42 & -8.35$\pm$0.31 & -9.14 \\
ResGen & -10.97$\pm$0.29 & -9.25$\pm$0.95 & -10.96$\pm$0.42 & -11.75$\pm$0.42 & -9.41$\pm$0.23 & -10.34$\pm$0.39 & -8.74$\pm$0.24 & -10.20 \\
DST & -10.95$\pm$0.57 & -10.67$\pm$0.24 & -10.54$\pm$0.22 & -10.88$\pm$0.37 & -9.71$\pm$0.19 & -10.03$\pm$0.36 & -8.33$\pm$0.41 & -10.16 \\
GraphGA & -10.03$\pm$0.41 & -9.89$\pm$0.25 & -9.94$\pm$0.15 & -10.22$\pm$0.39 & -9.32$\pm$0.51 & -9.29$\pm$0.20 & -7.75$\pm$0.32 & -9.49 \\
MIMOSA & -10.96$\pm$0.57 & -10.69$\pm$0.24 & -10.51$\pm$0.23 & -10.81$\pm$0.39 & -9.66$\pm$0.25 & -10.02$\pm$0.36 & -8.33$\pm$0.41 & -10.14 \\
MolDQN & -6.73$\pm$0.12 & -6.51$\pm$0.15 & -7.09$\pm$0.16 & -6.79$\pm$0.26 & -5.92$\pm$0.26 & -6.27$\pm$0.10 & -6.87$\pm$0.20 & -6.60 \\
Pasithea & -10.86$\pm$0.29 & -10.31$\pm$0.09 & -10.69$\pm$0.27 & -10.92$\pm$0.35 & -9.69$\pm$0.32 & -9.77$\pm$0.21 & -8.06$\pm$0.22 & -10.04 \\
REINVENT & -9.87$\pm$0.31 & -9.48$\pm$0.39 & -9.61$\pm$0.36 & -9.69$\pm$0.29 & -8.70$\pm$0.25 & -8.92$\pm$0.38 & -7.25$\pm$0.21 & -9.07 \\
SCREENING & -10.86$\pm$0.26 & -10.90$\pm$0.54 & -10.73$\pm$0.45 & -10.86$\pm$0.22 & -9.80$\pm$0.23 & -9.91$\pm$0.30 & -8.15$\pm$0.26 & -10.17 \\
SELFIES-VAE-BO & -10.15$\pm$0.60 & -9.76$\pm$0.12 & -9.99$\pm$0.28 & -10.00$\pm$0.23 & -9.02$\pm$0.33 & -9.18$\pm$0.39 & -7.75$\pm$0.22 & -9.41 \\
SMILES GA & -9.56$\pm$0.17 & -9.56$\pm$0.37 & -10.00$\pm$0.26 & -9.61$\pm$0.19 & -8.80$\pm$0.20 & -9.21$\pm$0.23 & -7.54$\pm$0.32 & -9.18 \\
SMILES LSTM HC & -10.38$\pm$0.21 & -10.30$\pm$0.15 & -10.19$\pm$0.12 & -10.49$\pm$0.49 & -9.36$\pm$0.17 & -9.71$\pm$0.43 & -7.90$\pm$0.26 & -9.76 \\
SMILES-VAE-BO & -9.93$\pm$0.22 & -9.78$\pm$0.10 & -9.96$\pm$0.29 & -10.05$\pm$0.20 & -9.03$\pm$0.30 & -9.18$\pm$0.39 & -7.74$\pm$0.25 & -9.38 \\
\hline
\ourM{} & -12.33$\pm$0.11 & -12.26$\pm$0.18 & -11.90$\pm$0.32 & -12.40$\pm$0.22 & -11.85$\pm$0.19 & -10.86$\pm$0.22 & -8.97$\pm$0.25 & -11.51 \\
\ourM{} + EFO & -12.83$\pm$0.26 & -12.76$\pm$0.22 & -12.31$\pm$0.19 & -12.49$\pm$0.28 & -11.88$\pm$0.20 & -11.08$\pm$0.24 & -9.07$\pm$0.31 & \textbf{-11.77} \\
\hline
\end{tabular}}
\caption{Top-10 average docking scores on additional benchmark(lower is better).}
\label{tab:sbdd7}
\end{table*}

This task aims to generate small-molecule ligands that bind favorably to a given protein pocket under fixed receptor geometry.

\paragraph{Data.}
Following standard practice in pocket-aware molecular generation~\cite{peng2022pocket2mol,guan20233d}, we use the CrossDocked2020 dataset~\cite{francoeur2020three}.
The processed dataset contains approximately 100,000 protein pocket--ligand complexes for training, with 100 target protein pockets held out for evaluation.

\paragraph{Baselines.}
We compare \ourM{} against representative structure-conditioned generation and optimization methods evaluated under the same protocol, including TargetDiff~\cite{guan20233d}, Pocket2Mol~\cite{peng2022pocket2mol}, DecompDiff~\cite{guan2024decompdiff}, MolCRAFT~\cite{qu2024molcraft}, RGA+Vina~\cite{fu2022reinforced}, DecompOpt~\cite{zhou2024decompopt}, and MOLCHORD~\cite{zhang2025molchord}.

\paragraph{Training and Inference Protocol.} We adopt a two-stage training strategy: supervised fine-tuning followed by Step-PPO maximization of Eq.~\ref{eq:reward struct}. We set $S_{\mathrm{ref}}=-9.0$, where the quadratic term implicitly shifts optimization focus to QED and SA as affinity improves. To balance the magnitude disparity between Vina scores and property metrics, we set $\lambda_1=7/3$ and $\lambda_2=5/6$. EFO is excluded in this benchmark to ensure fair comparison. See Appendix~\ref{app:training_details} for details.

\paragraph{Evaluation Metrics.}
We evaluate 100 molecules generated for each pocket using the following metrics:
(1) \textbf{Vina Dock}: Binding affinity calculated by AutoDock Vina~\cite{eberhardt2021autodock} under the protocol of~\cite{guan20233d};
(2) \textbf{QED}: Quantitative Estimate of Drug-likeness~\cite{bickerton2012quantifying};
(3) \textbf{SA}: Synthetic Accessibility score~\cite{ertl2009estimation};
and (4) \textbf{Diversity}: The average pairwise Tanimoto distance among generated molecules per pocket.
Following prior benchmarks~\cite{long2022zero,guan2024decompdiff}, we also report the \textbf{Success Rate}, defined as the percentage of valid molecules simultaneously satisfying
$\mathrm{Vina} < -8.18$, $\mathrm{QED} > 0.25$, and $\mathrm{SA} > 0.59$.
(5) \textbf{High Affinity}: Following prior evaluation protocols, we compute High Affinity as the percentage of generated molecules whose docking scores are no worse than those of the test-set ligands.

\paragraph{Results.}
Table~\ref{tab:res1} shows that \ourM{} establishes a new state-of-the-art with a \textbf{69.7\% Success Rate}, surpassing the best baseline by over 16\%. 
Unlike methods that sacrifice molecular quality for raw docking scores, \ourM{} achieves a superior balance, dominating in \textbf{QED} and \textbf{SA} while maintaining strong affinity. 
Notably, it also retains the highest \textbf{Diversity}, demonstrating that Step-PPO effectively optimizes binding without suffering from the mode collapse typically associated with RL. (See Appendix \ref{app:fig} for visual examples of generated molecules).
We also conducted experiments on an additional benchmark following \cite{zheng2024structure} to further demonstrate robustness (see Appendix~\ref{app:additional_experiments}).

\begin{figure*}[t]
    \centering
    \includegraphics[width=0.90\textwidth]{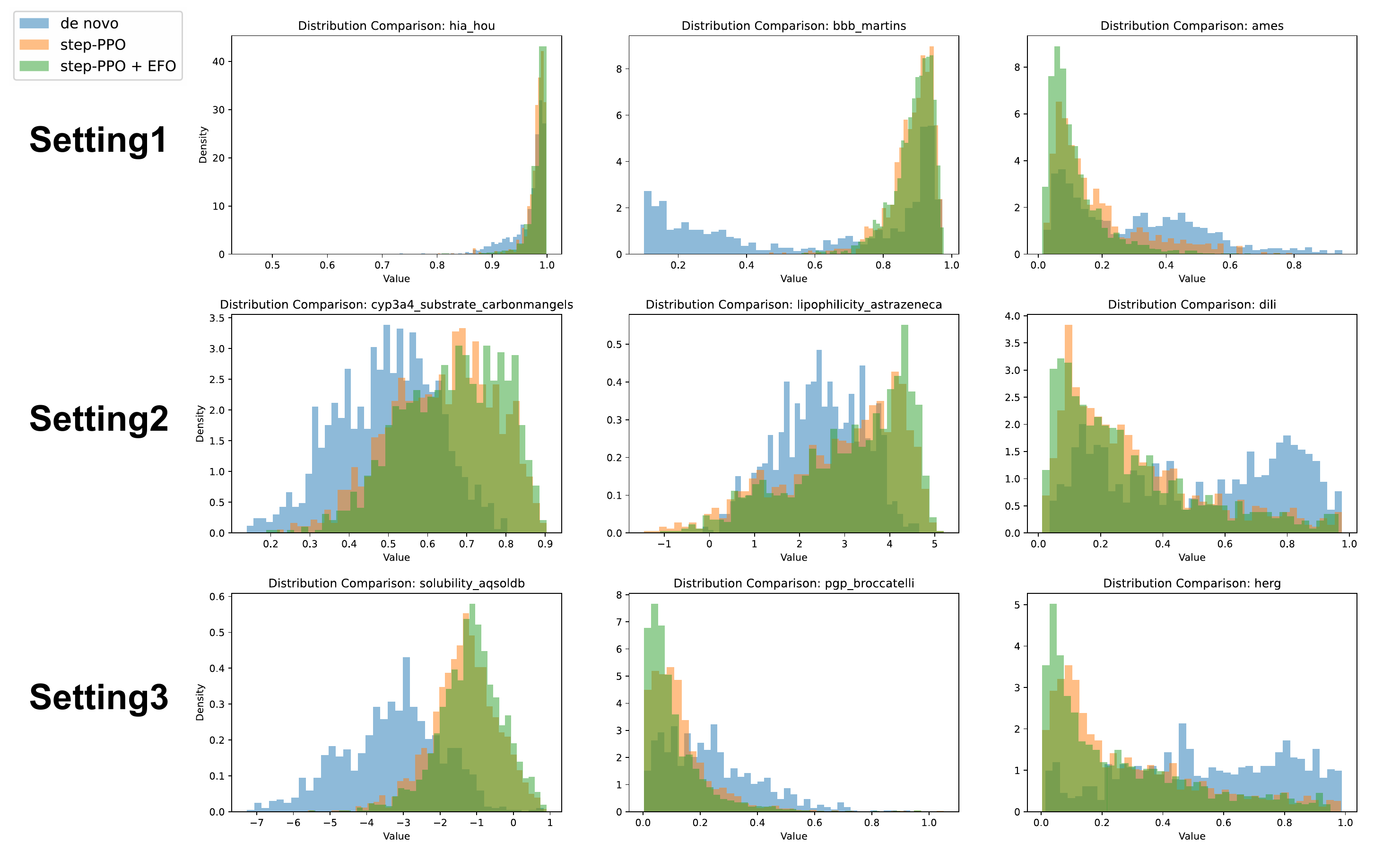}
    \caption{Histograms of Distribution shift under ADMET constraints for three settings.}
    \label{fig:admet_shift}
\end{figure*}

\begin{table*}[htbp]
\centering
\begin{tabular}{llcccc}
\hline
\textbf{Setting} & \textbf{Property} & \textbf{De novo} & \textbf{Step-PPO} & \textbf{Step-PPO + EFO} & \textbf{Target} \\
\hline
\multirow{3}{*}{Setting 1} 
 & HIA        & 0.97 & 0.98 & 0.99 & $\uparrow$ (1) \\
 & BBB        & 0.53 & 0.89 & 0.89 & $\uparrow$ (1) \\
 & Ames       & 0.37 & 0.21 & 0.15 & $\downarrow$ (0) \\
\hline
\multirow{3}{*}{Setting 2} 
 & CYP3A4sub  & 0.48 & 0.64 & 0.69 & $\uparrow$ (1) \\
 & LogP       & 2.3  & 3.4  & 3.5  & [3,5] \\
 & DILI       & 0.55 & 0.30 & 0.29 & $\downarrow$ (0) \\
\hline
\multirow{3}{*}{Setting 3} 
 & Solubility & -3.7 & -1.2 & -1.1 & $> -1$ \\
 & hERG       & 0.60 & 0.35 & 0.32 & $\downarrow$ (0) \\
 & Pgp\_Sub   & 0.22 & 0.16 & 0.12 & $\downarrow$ (0) \\
\hline
\end{tabular}
\caption{ADMET evaluation results under different settings.}
\end{table*}

\subsection{Property-Conditioned Generation}
\label{sec:exp_admet}

We evaluate intrinsic property conditioning on three practically motivated ADMET settings. We use MiniMol \cite{klaser2024texttt} as the property predictor to provide supervision and reward signals.

\paragraph{ADMET settings.}
We consider three multi-constraint targets:
\textbf{Setting 1 (CNS drugs):}
$\mathrm{HIA}=1$, $\mathrm{BBB}=1$, $\mathrm{Ames}=0$.
\textbf{Setting 2 (Hepatically metabolized drugs):}
$\mathrm{CYP3A4}_{\mathrm{sub}}=1$, $\mathrm{DILI}=0$, $\mathrm{LogP} \in [3,5]$.
\textbf{Setting 3 (Peripheral drugs):}
$\mathrm{Solubility} > -1$, $\mathrm{hERG}=0$, $\mathrm{Pgp\_Sub}=0$.

\paragraph{Training and Inference.}
For each setting, we first sample 10,000 \textit{de novo} generated molecules from the unconditional base model, and annotate each molecule using MiniMol.
For binary properties, we convert predicted probabilities into hard labels $\{0,1\}$ to form a pseudo-labeled training set, and then train \ourM{} with supervised learning under the corresponding target property vector.

And then, we further optimize \ourM{} using Step-PPO where the terminal reward is computed from MiniMol-predicted probabilities (without hard-thresholding) to preserve gradient-free but smooth optimization signals.
Finally, due to computational constraints, we run EFO for three generations to refine the candidate pool. Detailed hyperparameters and computational costs are provided in Appendix~\ref{app:training_details}.

\paragraph{Results.}
We visualize the distribution shift of each target property via histograms in Figure \ref{fig:admet_shift} across three distinct stages:
(i) Unconditional Generation from the base model;
(ii) Step-PPO Optimized; and
(iii) Step-PPO + EFO, where the candidates are further refined during inference.
This progression highlights how Step-PPO effectively shifts the molecular distribution toward the desired properties, while EFO provides a final sharpening of constraint satisfaction.

\subsection{Dual-Conditioned Generation}
\label{sec:exp_dual}

In real-world drug discovery, candidate molecules are required to simultaneously achieve strong binding affinity to the target protein and favorable ADMET properties. To evaluate the capability of \ourM{} under such realistic constraints, we conduct a dual-conditioned generation experiment that jointly enforces structure-based binding and toxicity-related property requirements.

We consider the protein 3o96\_A, an important therapeutic target in the PI3K–AKT signaling pathway. For this target, we condition molecule generation on both the fixed protein pocket and a safety constraint requiring Ames-negative predictions. Using the same training and optimization pipeline as in previous experiments, we generate 100 candidate molecules and evaluate them in terms of docking affinity and predicted Ames toxicity.

Results in Table \ref{tab:dual_res} show that \ourM{} achieves the best overall performance under both criteria, producing molecules with superior docking scores while maintaining the highest proportion of Ames-negative candidates among all compared methods. This demonstrates that \ourM{} can effectively balance binding optimization and toxicity avoidance within a unified framework, highlighting its practical value for realistic drug design scenarios. Detailed experimental settings and additional analyzes are provided in Appendix~\ref{app:dual}.

\begin{table}[h]
\centering
\resizebox{\linewidth}{!}{
\begin{tabular}{lcccc}
\hline
Methods & Vina Dock ($\downarrow$) & QED ($\uparrow$) & SA ($\uparrow$) & Ames ($\downarrow$) \\
\hline
Pocket2Mol  & -10.38 & 0.71 & 0.70 & 0.54 \\
TargetDiff  & -10.80 & 0.39 & 0.51 & 0.44 \\
FLAG        & -6.38  & 0.60 & 0.67 & 0.36 \\
MolCRAFT    & \underline{-11.33} & 0.43 & 0.66 & 0.49 \\
DecompDiff  & \textbf{-12.33} & 0.26 & 0.54 & 0.57 \\
\hline
w/o Ames Condition & -10.13 & \underline{0.83} & \textbf{0.88} & \underline{0.34} \\
\ourM{}     & -9.94 & \textbf{0.84} & \textbf{0.88} & \textbf{0.18} \\
\hline
\end{tabular}
}
\caption{Comparison under dual-conditioned generation on the 3o96\_A pocket.}
\label{tab:dual_res}
\end{table}

\section{Conclusion}
We propose \ourM{}, a unified framework for goal-directed molecular generation. By effectively handling structural, property, and dual constraints, \ourM{} achieves state-of-the-art performance across diverse benchmarks while maintaining quality and diversity, demonstrating strong potential for practical drug discovery applications.

\section*{Limitations}
Despite the improvements of \ourM{}, several limitations remain. Limited computational resources prevented large-scale training and extensive benchmarking, forcing us to initialize the model with pretrained weights rather than training from scratch. Additionally, the reliance on predictions from tools and models like ESM-2, AutoDock Vina and MiniMol, without accounting for potential prediction errors, may impact overall performance.

\bibliography{custom}

\appendix
\section{Case Study}
\begin{figure*}[t]
    \centering
    \includegraphics[width=\textwidth]{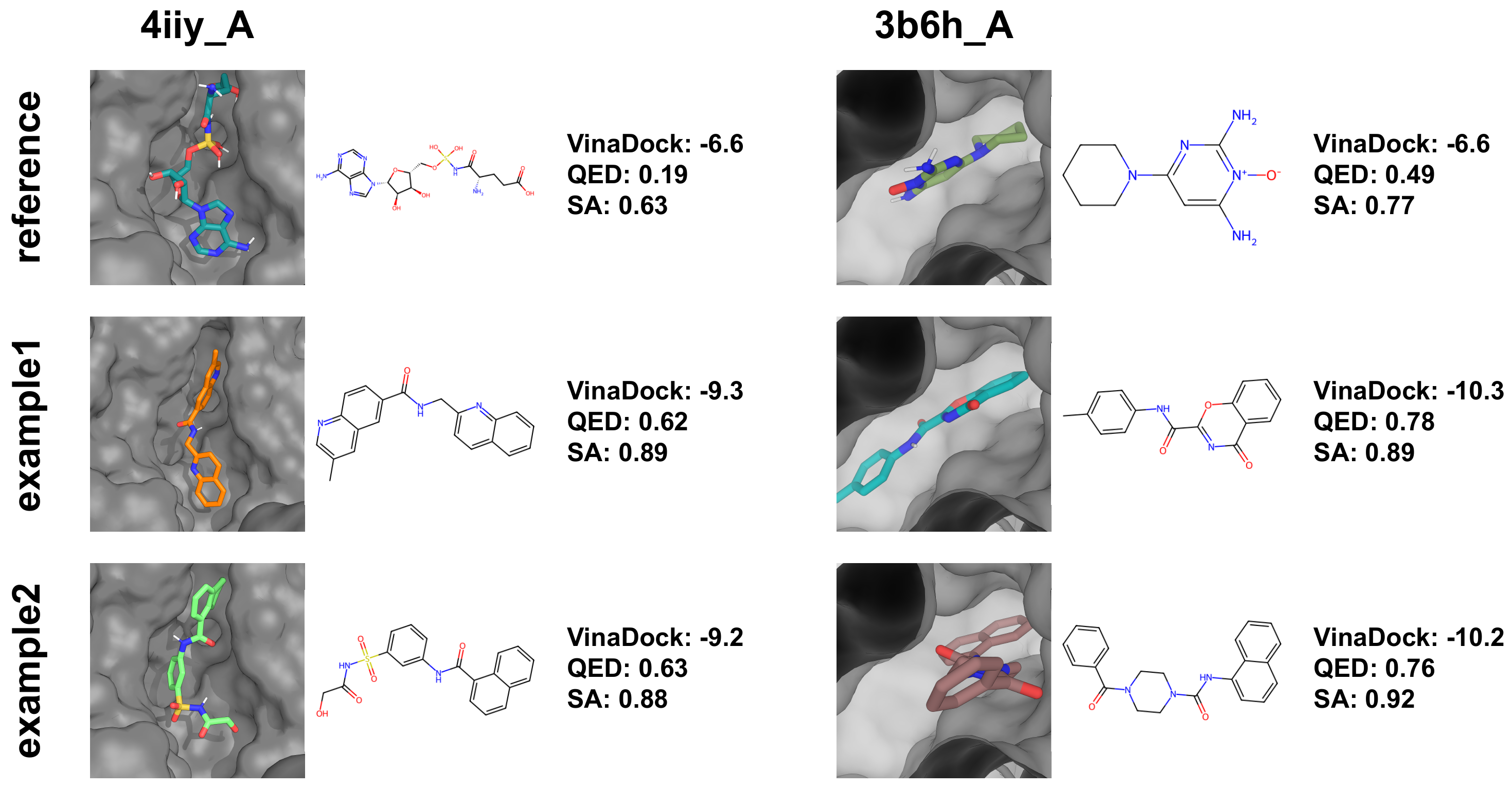}
    \caption{Case Study. Two pockets are shown. For each pocket, we visualize the reference ligand and two \ourM{}-generated ligands with their docking poses in the pocket.}
    \label{fig:case1}
\end{figure*}

Figure~\ref{fig:case1} visualizes two representative pockets from the benchmark. For each pocket, we show the reference ligand from the dataset and two examples generated by \ourM{} with their binding poses, 2d graph and metrics. Results show that our method generate better molecules in every metric.
\label{app:fig}

\section{Residue-level Physicochemical Feature Definition}
\label{app:phys_feature}

To explicitly encode surface biochemical properties of protein pockets, we construct a residue-level physicochemical feature representation based on amino acid identity. For each residue, we define a five-dimensional feature vector capturing hydrophobicity, electrostatic charge, polarity, and hydrogen-bonding capability. These features are computed deterministically and do not require additional learning.

\paragraph{Hydropathy.}
We adopt the Kyte–Doolittle hydropathy index for each amino acid, which quantifies residue hydrophobicity on a continuous scale. To ensure numerical stability and compatibility with neural network inputs, the raw hydropathy values are normalized by a factor of 5. For example, isoleucine (I) has a value of 4.5, while arginine (R) has a value of -4.5.

\paragraph{Electrostatic Charge.}
Residue charge is encoded as a discrete scalar. Positively charged residues arginine (R) and lysine (K) are assigned a value of +1, negatively charged residues aspartic acid (D) and glutamic acid (E) are assigned -1, and histidine (H) is assigned a partial charge of +0.1 to reflect its conditional protonation state. All other residues are assigned 0.

\paragraph{Polarity.}
Polarity is represented as a binary indicator. Polar residues (R, N, D, Q, E, H, K, S, T, Y) are assigned 1, while nonpolar residues are assigned 0.

\paragraph{Hydrogen Bond Acceptor.}
A binary feature indicates whether a residue can act as a hydrogen bond acceptor. Residues capable of accepting protons (D, E, N, Q, H, S, T, Y) are assigned 1, and all others are assigned 0.

\paragraph{Hydrogen Bond Donor.}
Similarly, hydrogen bond donor capability is encoded as a binary indicator. Residues that can donate protons (R, K, W, N, Q, H, S, T, Y) are assigned 1, while the remaining residues are assigned 0.

This deterministic encoding provides an explicit and interpretable description of residue surface chemistry, complementing the learned semantic representations extracted from protein language models.

\section{Structured Context Segment of Conditions}
\label{app:context segment}
To enable conditional generation, we augment the input sequence with a structured context segment that encodes the conditioning signal. Rather than injecting a raw latent vector, the context is wrapped with dedicated special tokens to explicitly indicate its semantic boundaries and type.

Concretely, the context segment is constructed as
\[
\langle\texttt{boc}\rangle\,
\langle\texttt{boe}\rangle\, \mathbf{h}_{ext}\, \langle\texttt{eoe}\rangle\,
\langle\texttt{boi}\rangle\, \mathbf{h}_{int}\, \langle\texttt{eoi}\rangle\,
\langle\texttt{eoc}\rangle,
\]
where $\mathbf{h}_{ext}$ and $\mathbf{h}_{int}$ denote the extrinsic (structural) and intrinsic (property) condition embeddings produced by the Unified Constraint Adaptor, respectively. The special tokens
$\langle\texttt{boc}\rangle / \langle\texttt{eoc}\rangle$ mark the beginning and end of the entire context segment, while
$\langle\texttt{boe}\rangle / \langle\texttt{eoe}\rangle$ and
$\langle\texttt{boi}\rangle / \langle\texttt{eoi}\rangle$
delimit the extrinsic and intrinsic condition blocks. All special tokens are associated with learnable word embeddings and are processed identically to molecular tokens within the Transformer.

\section{Implementation Details}
\label{app:training_details}

Our training pipeline consists of two phases: supervised fine-tuning and reinforcement learning optimization.

\textbf{Model Architecture.}
Our model is based on a BERT-style architecture with around 100 million parameters, comprising 12 Transformer encoder layers. Each layer employs multi-head self-attention and feed-forward sublayers, consistent with the original BERT design.

\textbf{Supervised Fine-Tuning.} We first adapt the unconditional GenMol backbone to the structure-conditioned task. The model is fine-tuned end-to-end on the CrossDocked2020 training set using the \textbf{AdamW} optimizer. We set the learning rate to $5\times10^{-5}$ and the batch size to 256. Training is conducted for 10 epochs in precision bf16, which requires approximately 1.5 hours on a single NVIDIA A800 GPU.

\textbf{Step-PPO Optimization.} Starting from the supervised checkpoint, we further optimize the model via Step-PPO to maximize the composite reward defined in Eq.~\ref{eq:reward struct}.
\begin{itemize}\item \textbf{Reward Configuration:} We set the reference affinity barrier to $S_{\mathrm{ref}}=-9$, which is stricter than the evaluation success threshold of $-8.18$. The QED and SA components are computed using the TDC oracle~\cite{huang2021therapeutics}.\item \textbf{Hyperparameters:} We use a learning rate of $1\times10^{-5}$ and a batch size of 128. The PPO clipping parameter is set to $\epsilon=0.2$, and the entropy regularization coefficient is set to $\beta=0.01$. Optimization runs for up to 150 steps, with 2 epochs per step.\item \textbf{Sampling \& Stopping:} During rollouts, we use a sampling temperature of 0.5. We employ an early stopping mechanism that terminates training when the batch success rate exceeds 80\%.\item \textbf{Compute Cost:} Due to the computational expense of on-policy docking evaluations, this phase requires approximately 190 GPU-hours.\end{itemize}

\textbf{Inference Settings.} For the CrossDocked2020 benchmark comparison, we generate molecules using the Step-PPO fine-tuned model without the Evolutionary Fragment Optimization (EFO) module to strictly adhere to the generation budget of baseline methods.

\section{Details of Supervised Learning}
\label{app:obj}
We first train \ourM{} in a supervised manner to provide a stable initialization for subsequent optimization stages. Here we will continue using the symbols from Section~\ref{sec:background_mdlm}.

Given a conditioning context $\mathbf{c}$, the denoising network is trained to predict the original molecular tokens from the masked sequence.
Since the context $\mathbf{c}$ serves purely as external guidance and is not part of the generation target, no loss is applied to the condition tokens during supervised training. Instead, the objective is defined solely over the molecular sequence.

Following MDLM~\cite{sahoo2024simple}, we optimize a continuous-time approximation of the negative evidence lower bound (NELBO):
\begin{equation}
    \resizebox{1\hsize}{!}{$
    \mathcal{L}_{\mathrm{NELBO}}
    =
    \mathbb{E}_{q}
    \int_{0}^{1}
    \frac{\alpha'_t}{1-\alpha_t}
    \sum_{l=1}^{L}
    \log
    \left\langle
    \pmb{x}^{l}_\theta(\pmb{z}_t, t, \mathbf{c}),
    \pmb{x}^{l}
    \right\rangle
    \,\mathrm{d}t,
    $}
    \label{eq:obj_masked}
\end{equation}
where $\pmb{x}^{l}_\theta(\pmb{z}_t, t, \mathbf{c})$ denotes the predicted categorical distribution for the $l$-th molecular token conditioned on the noisy sequence $\pmb{z}_t$, diffusion time $t$, and the context $\mathbf{c}$. This objective corresponds to a time-weighted masked language modeling loss over molecular tokens.

This supervised stage adapts the pre-trained unconditional backbone to operate in the presence of conditioning signals, enabling the model to incorporate contextual information into the denoising process while retaining its original diffusion formulation. It therefore establishes a condition-aware initialization for subsequent optimization.

\section{Details of Evolutionary Fragment Optimization}
\label{app:de_efo}

Evolutionary Fragment Optimization (EFO) is an iterative evolutionary procedure that integrates fragment-level exploration with masked discrete diffusion.
Unlike classical genetic algorithms \cite{jensen2019graph, fu2021differentiable}that rely on random atomic mutations, EFO operates on a dynamic vocabulary of chemically meaningful fragments and employs a structured remasking strategy, enabling efficient traversal of chemical space while optimizing task-specific objectives.
The optimization process is centered around a fragment vocabulary $\mathcal{V}$, which serves as a genetic pool for molecule construction and evolution.

To prioritize high-value substructures, we define a scoring function $S(f_k)$ for each fragment $f_k$ extracted from a source dataset $\mathcal{D}$.
The score reflects the average target property value of molecules containing the fragment:
\begin{equation}
S(f_k) = \frac{1}{|\mathcal{S}(f_k)|} \sum_{x \in \mathcal{S}(f_k)} y(x),
\label{eq:efo1}
\end{equation}
where $\mathcal{S}(f_k) = \{x \in \mathcal{D} : f_k \text{ is a subgraph of } x\}$.
The initial vocabulary $\mathcal{V}$ is constructed by selecting the top-$V$ fragments ranked by $S(f_k)$ and is dynamically updated throughout the optimization process to incorporate newly discovered high-scoring fragments.

The EFO procedure iterates over a generative cycle consisting of four tightly integrated stages: initialization, mutation, guided reconstruction, and vocabulary evolution.

\begin{enumerate}
    \item \textbf{Initialization via Fragment Attachment.}
    Each iteration begins by constructing a seed molecule $x_{\mathrm{init}}$.
    Two fragments are randomly sampled from the current vocabulary $\mathcal{V}$ and attached to form a valid Sequential Attachment-based Fragment Embedding (SAFE) representation.
    This initialization strategy ensures that the starting molecules already contain substructures statistically correlated with favorable target properties.

    \item \textbf{Mutation via Fragment Remasking.}
    To explore the local chemical neighborhood of $x_{\mathrm{init}}$, we apply a mutation operator termed \emph{Fragment Remasking}.
    Unlike token-level masking, this operator acts at the semantic level of chemical substructures.
    A fragment is selected according to a decomposition rule $\mathcal{R}_{\mathrm{remask}}$ and replaced by a sequence of mask tokens $[M]$.
    The number of mask tokens $m$ is sampled from a predefined distribution $p_{\mathrm{len}}$, such as the empirical fragment-length distribution observed in the training data.
    This mechanism allows flexible control over the size and complexity of the regenerated substructure.
    Conceptually, this operation corresponds to Gibbs sampling, where a fragment $f_k$ is resampled from the conditional distribution $p(f_k \mid f_{\setminus k})$, with $f_{\setminus k}$ denoting the unmasked molecular context.

    \item \textbf{Reconstruction with Molecular Fragment Context.}
The masked region is reconstructed using the discrete diffusion model conditioned on the remaining molecular fragments.
Given a partially masked molecule, the diffusion model iteratively denoises the masked positions while attending to the unmasked fragment-level context through self-attention.
This conditional reconstruction naturally enforces chemical compatibility between the regenerated fragment and the existing molecular scaffold, as the denoising distribution is explicitly conditioned on the surrounding fragments.

Formally, at each diffusion step $t$, the model predicts the categorical distribution for selected masked token $x^l$ as
\begin{equation}
x_{\theta,i}^l(z_t, t) = p_\theta(x^l = i \mid z_t),
\end{equation}
where $z_t$ denotes the noisy sequence retaining the unmasked fragment context.
By sampling from this conditional distribution across diffusion steps, the model generates fragments that are coherent with the molecular structure and aligned with the learned chemical priors.
This context-aware reconstruction serves as a structured mutation operator, enabling localized yet chemically valid exploration of the molecular space.

    \item \textbf{Vocabulary Evolution.}
    The newly generated molecule $x_{\mathrm{new}}$ is evaluated by the task-specific scoring oracle.
    It is then decomposed into fragments, which are scored using $S(\cdot)$ and merged into the vocabulary.
    The vocabulary $\mathcal{V}$ is subsequently updated by retaining the top-$V$ fragments from the union of the existing and newly generated candidates.
    This feedback loop enables continual expansion and refinement of the fragment pool, progressively steering the search toward regions of chemical space associated with higher target property values.
\end{enumerate}

\begin{algorithm}[t]
\small
\caption{Evolutionary Fragment Optimization (EFO)}
\label{alg:efo}
\begin{algorithmic}
\STATE \textbf{Input:}
Dataset of molecules $\mathcal{D}$;
vocabulary size $V$;
fragment decomposition rule $\mathcal{R}_{\mathrm{vocab}}$;
fragment remasking rule $\mathcal{R}_{\mathrm{remask}}$;
number of generations $G$

\STATE Decompose $\mathcal{D}$ into a fragment multiset $\mathcal{F}$ using $\mathcal{R}_{\mathrm{vocab}}$
\STATE Initialize fragment vocabulary $\mathcal{V}$ with the top-$V$ fragments from $\mathcal{F}$ ranked by Eq.~(\ref{eq:efo1})
\STATE Estimate fragment-length distribution $p_{\mathrm{len}}$ from $\mathcal{D}$ based on $\mathcal{R}_{\mathrm{remask}}$
\STATE Initialize generated molecule set $\mathcal{M} \gets \emptyset$

\WHILE{$|\mathcal{M}| < G$}
    \STATE Sample two fragments from $\mathcal{V}$ and attach them to form an initial molecule $x_{\mathrm{init}}$
    \STATE Sample mask length $m \sim p_{\mathrm{len}}$
    \STATE Select a fragment in $x_{\mathrm{init}}$ according to $\mathcal{R}_{\mathrm{remask}}$
    \STATE Replace the selected fragment with $m$ mask tokens to obtain a partially masked molecule $x_{\mathrm{mask}}$
    \STATE Reconstruct the masked region via conditional discrete diffusion to obtain $x_{\mathrm{new}}$
    \STATE Update $\mathcal{M} \gets \mathcal{M} \cup \{x_{\mathrm{new}}\}$
    \STATE Decompose $x_{\mathrm{new}}$ into fragments $\{f_1, f_2, \dots\}$ using $\mathcal{R}_{\mathrm{vocab}}$
    \STATE Update $\mathcal{V}$ by retaining the top-$V$ fragments from $\mathcal{V} \cup \{f_1, f_2, \dots\}$
\ENDWHILE

\STATE \textbf{Output:} Generated molecule set $\mathcal{M}$
\end{algorithmic}
\end{algorithm}

\section{Details of Dual-Conditioned Generation}
\label{app:dual}

In this section, we provide detailed experimental settings and quantitative results for the dual-conditioned generation task discussed in Sec.~\ref{sec:exp_dual}. As noted in the main text, realistic drug discovery requires candidate molecules to simultaneously achieve strong target binding affinity and acceptable safety profiles. Optimizing binding affinity alone often leads to toxic or developability-limited compounds, while focusing solely on ADMET properties may result in insufficient target engagement. This experiment is designed to evaluate whether \ourM{} can effectively reconcile these heterogeneous and potentially competing objectives within a unified generation framework.

\paragraph{Target protein.}
We consider the protein structure with PDB ID \textbf{3O96}, corresponding to the N-terminal pleckstrin homology (PH) domain of human \textbf{AKT1}, a key serine/threonine kinase in the PI3K–AKT signaling pathway. AKT1 is ubiquitously expressed across multiple human tissues and plays a central role in regulating cell survival, metabolism, and proliferation. Dysregulation of AKT1 signaling is closely associated with cancer and metabolic disorders, making it an important therapeutic target.
Ligands targeting the AKT1 PH domain must be able to reach intracellular AKT1 while maintaining sufficient selectivity. Moreover, due to the essential physiological functions of AKT1 in normal organs, candidate compounds are required to exhibit low systemic toxicity and acceptable safety profiles, including the absence of mutagenic potential as indicated by a negative Ames test.

\paragraph{Experimental setup.}
We condition \ourM{} on both the fixed receptor structure of 3O96 and an intrinsic toxicity constraint requiring generated molecules to be \textbf{Ames-negative}. Training follows the same two-stage paradigm used throughout this work. Specifically, we first perform supervised conditional training, after which Step-PPO is applied to optimize a composite reward that jointly accounts for docking affinity (evaluated by AutoDock Vina) and MiniMol-predicted Ames toxicity probability. During inference, we generate \textbf{100 molecules} for the target pocket and evaluate each candidate under both structural and ADMET-related metrics.

To better isolate the effect of toxicity conditioning, we additionally report an ablation variant (\textbf{w/o Ames Condition}) in which \ourM{} is trained and optimized solely for structure-based binding without enforcing the Ames constraint.

\paragraph{Evaluation metrics.}
Generated molecules are evaluated using the following criteria:
(1) \textbf{Vina Dock}: Binding affinity to the AKT1 PH domain;
(2) \textbf{QED}: Quantitative estimate of drug-likeness;
(3) \textbf{SA}: Synthetic accessibility score;
(4) \textbf{Ames}: Predicted mutagenicity probability, where lower values indicate safer compounds.
This evaluation protocol reflects a realistic screening scenario in which candidates must simultaneously satisfy potency, developability, and safety requirements.

\paragraph{Results.}
Quantitative results are summarized in Table~\ref{tab:dual_res}. Compared with existing structure-conditioned baselines, \ourM{} achieves the lowest predicted Ames toxicity while maintaining competitive docking performance and substantially higher molecular quality in terms of QED and SA. Notably, methods that achieve very strong docking scores often suffer from significantly elevated Ames toxicity, highlighting the limitation of affinity-only optimization.

The comparison between \ourM{} and its ablation variant further demonstrates the effect of explicit toxicity conditioning. While the \textbf{w/o Ames Condition} model achieves slightly stronger docking scores, it exhibits nearly twice the Ames toxicity compared to the full model. In contrast, \ourM{} effectively reduces mutagenicity risk with only a marginal trade-off in binding affinity, resulting in a more balanced and practically viable candidate set.

Overall, these results confirm that \ourM{}, empowered by Step-PPO, can successfully coordinate structure-based optimization and safety constraints, avoiding the collapse to single-objective solutions and enabling realistic dual-conditioned molecular generation.

\section{Evaluation on Additional SBDD Benchmark}
\label{app:additional_experiments}

\paragraph{Benchmark.}
We further evaluate \ourM{} on the standardized structure-based drug design benchmark proposed in \cite{zheng2024structure}, which uses seven representative target proteins (PDBIDs: 1IEP, 3EML, 3NY8, 4RLU, 4UNN, 5MO4, 7L11) and evaluates top-10 docking performance across diverse algorithmic families.

\paragraph{Protocol.}
To test generalization across targets, we initialize \ourM{} using the model already fine-tuned on CrossDocked2020 (Sec.~\ref{sec:exp_struct}), and then perform additional Step-PPO optimization under each target pocket using the benchmark docking oracle. Following the benchmark setting, we report the average Top-10 docking score for each target and the overall average.
Since this benchmark setting naturally supports iterative black-box optimization, we using 3-round EFO at inference time.

\paragraph{Results.}
Table~\ref{tab:sbdd7} shows that \ourM{} is competitive with strong optimization-based baselines while maintaining a learned generative prior. We also observe that enabling EFO consistently improves the docking performance by further exploiting high-reward fragment motifs.

\section{Reward Analysis}
\label{app:training_dynamics}

To empirically validate the optimization stability and convergence speed of our Step-PPO algorithm, we visualize the reward trajectories during the alignment phase for the three property-conditioned generation tasks described in Section~\ref{sec:exp_admet}.

\begin{figure}[h]
    \centering
    \includegraphics[width=\linewidth]{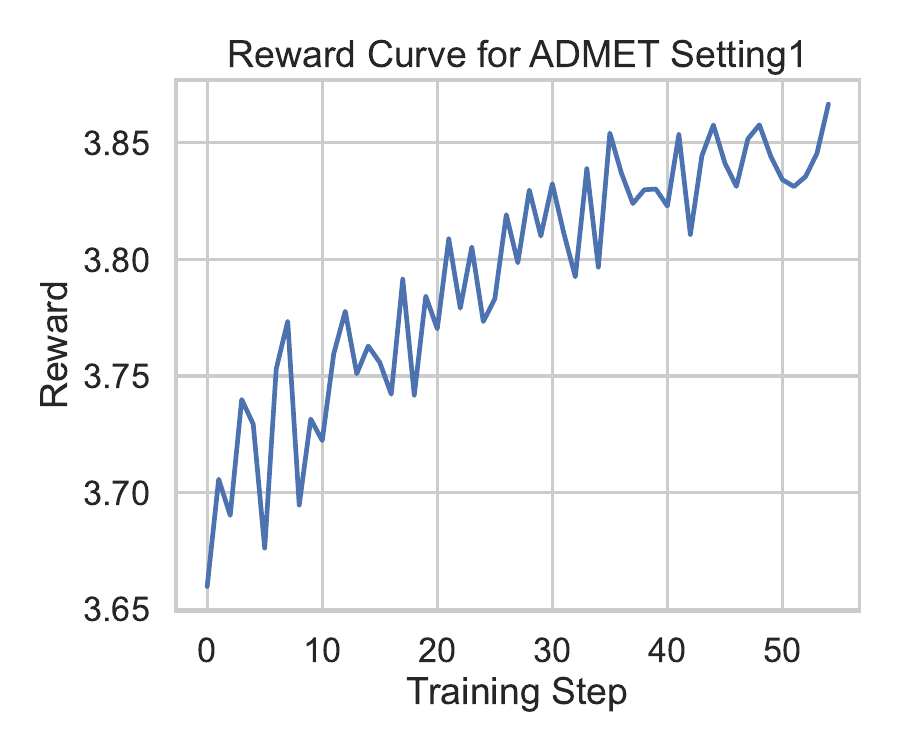}
    \caption{Setting 1 CNS Drugs}
    \label{fig:reward_cns}
\end{figure}

\begin{figure}[h]
    \centering
    \includegraphics[width=\linewidth]{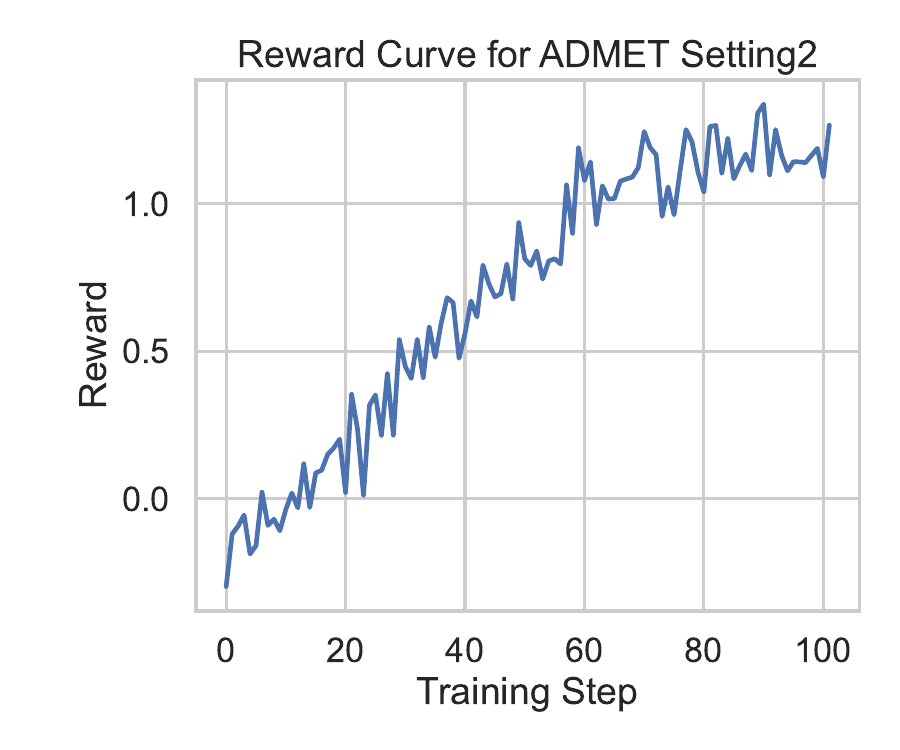} 
    \caption{Setting 2 Hepatic Drugs}
    \label{fig:reward_hepatic}
\end{figure}

\begin{figure}[h]
    \centering
    \includegraphics[width=\linewidth]{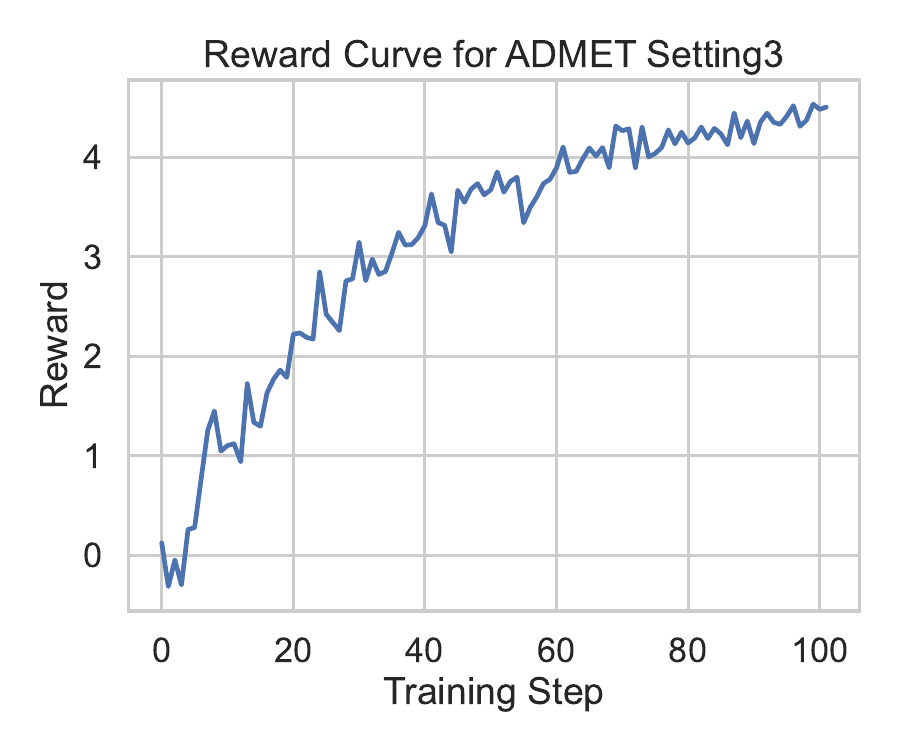}
    \caption{Setting 3 Peripheral Drugs}
    \label{fig:reward_peripheral}
\end{figure}

Figure~\ref{fig:reward_cns}-\ref{fig:reward_peripheral} illustrates the average reward curves throughout the Step-PPO training process. We observe consistent convergence patterns across all three distinct ADMET settings:

\begin{itemize}
    \item \textbf{Fast Convergence:} In all scenarios, the model effectively learns to navigate the chemical space towards high-reward regions within the first hundred steps. This rapid adaptation demonstrates that the supervised initialization provides a strong chemical prior, allowing Step-PPO to focus immediately on constraint satisfaction rather than relearning basic chemical validity.
    \item \textbf{Stability:} Unlike standard RL fine-tuning which often suffers from high variance or collapse, our step-wise formulation maintains a steady ascending trajectory. The relatively narrow variance (if applicable in your plot) suggests that the token-level policy updates are robust and do not degrade the overall structural integrity of the molecules.
    \item \textbf{Task Difficulty:} We note that Setting 2 (Hepatically metabolized drugs) shows a slightly slower convergence rate compared to Setting 1 and 3. This aligns with the complexity of the constraints, as satisfying specific Lipophilicity ranges combined with enzyme substrate specificity represents a more constrained optimization landscape.
\end{itemize}

These training dynamics confirm that \ourM{} can efficiently align the generative distribution with complex, multi-objective property constraints without requiring extensive hyperparameter tuning or suffering from mode collapse.

\section{Ablation Study}
\label{app:ablation}

In this section, we investigate the contribution of the specific architectural designs in the Unified Constraint Adaptor (UCA) and the impact of the reinforcement learning stage.

Note that the effectiveness of the Evolutionary Fragment Optimization (EFO) has already been demonstrated in Section~\ref{sec:exp_admet} and Appendix~\ref{app:additional_experiments}, where enabling EFO consistently improved performance. Therefore, we exclude EFO from this analysis.

We compare five variants to dissect the framework:
\begin{itemize}
    \item \textbf{SFT (Base):} A baseline supervised model where the UCA uses mean pooling instead of linear attention, and receives only ESM-2 embeddings without explicit physicochemical features.
    \item \textbf{SFT (w/o Attn.):} UCA uses mean pooling, but includes both ESM-2 and physicochemical features.
    \item \textbf{SFT (w/o Phys.):} UCA uses linear attention, but relies solely on ESM-2 embeddings (no physicochemical stream).
    \item \textbf{SFT (Full UCA):} The complete UCA architecture (Attention + Phys. Features) trained only with Supervised Fine-Tuning (no RL).
    \item \textbf{Full Model (Step-PPO):} The complete \ourM{} framework fine-tuned with Step-wise PPO.
\end{itemize}

The results on the CrossDocked2020 test set are summarized in Table~\ref{tab:ablation_components}.

\paragraph{Architecture Analysis (Impact of UCA Design).}
Comparing the SFT variants reveals the importance of our UCA design.
First, \textbf{SFT (w/o Phys.)} generally underperforms \textbf{SFT (Full UCA)}, indicating that while protein language models provide rich semantics, the explicit physicochemical features (charge, hydropathy, etc.) provide critical guidance for precise surface matching.
Second, replacing the linear attention with mean pooling (\textbf{SFT w/o Attn.}) leads to a performance drop. This suggests that the attention mechanism successfully learns to weigh critical residues in the pocket, whereas mean pooling dilutes the signal from key binding sites.
The \textbf{SFT (Base)} variant, lacking both designs, yields the lowest performance among the supervised models, validating the synergy of our dual-stream encoder and attention pooling.

\paragraph{Impact of Reinforcement Learning.}
A significant performance gap is observed between \textbf{SFT (Full UCA)} and the \textbf{Full Model (Step-PPO)}. While the supervised model with the full UCA architecture achieves reasonable validity and docking scores, it fails to reach the high-affinity regime. The introduction of Step-PPO drastically improves the Vina Dock score and Success Rate without collapsing diversity. This confirms that while the UCA provides a necessary condition-aware representation, the Step-wise PPO algorithm is indispensable for aligning the generation process with the stringent binding affinity objectives.

\paragraph{Extended Ablation: Role of Initialization, RL, and EFO.}
To further address the contribution of each component raised by the reviewer, we conduct additional controlled experiments on pocket 3o96\_A due to computational constraints. Specifically, we investigate whether the performance gains mainly stem from reinforcement learning (Step-PPO), and whether similar improvements can be achieved by applying Step-PPO or EFO on weaker initializations.

Table~\ref{tab:ablation_rl_init} shows the results. We compare: (i) the full model initialized from SFT (Full UCA), (ii) a weaker initialization SFT (Base) followed by Step-PPO, (iii) Step-PPO without any SFT initialization, and (iv) EFO-only variants.

\begin{table*}[h]
\centering
\begin{tabular}{lccc}
\hline
\textbf{Method} & \textbf{Vina Dock} ($\downarrow$) & \textbf{QED} ($\uparrow$) & \textbf{SA} ($\uparrow$) \\
\hline
SFT(Full)+Step-PPO & \textbf{-10.13} & 0.83 & 0.88 \\
SFT(Base)+Step-PPO & -9.97 & \textbf{0.84} & 0.88 \\
Step-PPO only & -9.49 & 0.84 & \textbf{0.89} \\
EFO only (3 iters) & -7.66 & 0.70 & 0.75 \\
EFO only (5 iters) & -8.02 & 0.69 & 0.77 \\
\hline
\end{tabular}
\caption{Ablation of initialization, RL fine-tuning, and EFO on pocket 3o96\_A.}
\label{tab:ablation_rl_init}
\end{table*}

These results provide several key insights. First, while Step-PPO significantly improves performance over supervised models, its effectiveness depends strongly on the initialization. Removing SFT leads to both degraded docking performance and substantially slower convergence, indicating that SFT provides a crucial chemical prior and stabilizes RL optimization. 

Second, applying Step-PPO on a weaker backbone (SFT Base) narrows but does not close the gap with the full model, suggesting that architectural improvements and RL contribute complementary gains. This also clarifies that our method is not equivalent to simply fine-tuning GenMol with RL, as the conditioning-aware UCA architecture remains essential.

Third, EFO alone yields limited improvements even with increased iterations, due to the extremely large combinatorial search space. This confirms that EFO acts as a refinement module rather than a standalone optimizer, and cannot replace the generative policy learned via RL.

Overall, these findings demonstrate that (1) SFT is critical for stable and efficient RL optimization, (2) Step-PPO is the primary driver for achieving high-affinity generation, and (3) EFO provides additional but limited gains through local refinement.

\section{Inference Efficiency and Runtime Analysis}
\label{app:efficiency}

In addition to generation performance, inference efficiency is an important practical factor for drug design.
Frag2Seq~\cite{fu2024fragment} reports the wall-clock time required to generate 100 molecules per pocket for a range of structure-conditioned molecular generation methods. Following the same evaluation protocol, we compare \ourM{} with these representative baselines.

We run on a single NVIDIA A800 GPU with a batch size of 100. We report the inference time for the base model as well as the variant augmented with EFO using 3 refinement rounds.

Table~\ref{tab:runtime} summarizes the runtime comparison. \ourM{} achieves orders-of-magnitude faster inference than diffusion-based and graph-based baselines. Even with EFO enabled, \ourM{} remains significantly faster than all compared methods, demonstrating its suitability for large-scale and iterative molecular generation scenarios.

\begin{table}[h]
\centering
\begin{tabular}{lc}
\hline
\textbf{Method} & \textbf{Time (s)} \\
\hline
3D-SBDD \cite{luo20213d} & 15986.4 \\
Pocket2Mol \cite{peng2022pocket2mol} & 2827.3 \\
GraphBP \cite{liu2022generating} & 1162.8 \\
TargetDiff \cite{guan20233d} & 3428.0 \\
DecompDiff \cite{guan2024decompdiff} & 6189.0 \\
DiffSBDD \cite{schneuing2024structure} & 629.9 \\
FLAG \cite{zhang2023molecule} & 1289.1 \\
DrugGPS \cite{zhang2023learning} & 1007.8 \\
Lingo3DMol \cite{feng2024generation} & 1481.9 \\
Frag2Seq \cite{fu2024fragment} & 48.8 \\
\hline
\ourM{} & \textbf{3.5} \\
\ourM{} + EFO (3 rounds) & \textbf{29.9} \\
\hline
\end{tabular}
\caption{Inference time (seconds) for generating 100 molecules per pocket. Lower is better.}
\label{tab:runtime}
\end{table}
\end{document}